\def\year{2021}\relax
\documentclass[letterpaper]{article} % DO NOT CHANGE THIS
\usepackage{aaai21}  % DO NOT CHANGE THIS
\usepackage{times}  % DO NOT CHANGE THIS
\usepackage{helvet} % DO NOT CHANGE THIS
\usepackage{courier}  % DO NOT CHANGE THIS
\usepackage[hyphens]{url}  % DO NOT CHANGE THIS
\usepackage{graphicx} % DO NOT CHANGE THIS
\usepackage{graphicx}  % DO NOT CHANGE THIS
\usepackage{natbib}  % DO NOT CHANGE THIS AND DO NOT ADD ANY OPTIONS TO IT
\usepackage{caption} % DO NOT CHANGE THIS AND DO NOT ADD ANY OPTIONS TO IT

\usepackage{booktabs}
\usepackage{amsfonts,amssymb}
\usepackage{multirow}
\usepackage{xspace}
\usepackage{hyperref}
\usepackage{subfigure}
\usepackage{bm}
\usepackage{color}
\usepackage{amsmath}
\usepackage{stmaryrd}
\usepackage{stfloats}

\newcommand{\ie}{\emph{i.e.,}\xspace}
\newcommand{\eg}{\emph{e.g.,}\xspace}

\title{GDPNet: Refining Latent Multi-View Graph for Relation Extraction}
\author {
        Fuzhao Xue\textsuperscript{\rm 1}, 
        Aixin Sun\textsuperscript{\rm 1}, 
        Hao Zhang\textsuperscript{\rm 1,2}, 
        Eng Siong Chng\textsuperscript{\rm 1} \\
}
\affiliations {
    \textsuperscript{\rm 1} School of Computer Science and Engineering, Nanyang Technological University, Singapore \\
    \textsuperscript{\rm 2} Institute of High Performance Computing, A*STAR, Singapore \\
    \{fuzhao001@e., axsun@, hao007@e., aseschng@\}ntu.edu.sg
}
\begin{document}

\maketitle
\begin{abstract}

Relation Extraction (RE) is to predict the relation type of two entities that are mentioned in a piece of text, \eg a sentence or a dialogue. When the given text is long, it is challenging to identify indicative words for the relation prediction. Recent advances on RE task are from BERT-based sequence modeling and graph-based modeling of relationships among the tokens in the sequence. In this paper, we propose to construct a latent multi-view graph to capture various possible relationships among tokens. We then refine this graph to select important words for relation prediction. Finally, the representation of the refined graph and the BERT-based sequence representation are concatenated for relation extraction. Specifically, in our proposed GDPNet (Gaussian Dynamic Time Warping Pooling Net), we utilize Gaussian Graph Generator (GGG) to generate edges of the multi-view graph. The graph is then refined by Dynamic Time Warping Pooling (DTWPool). On DialogRE and TACRED, we show that GDPNet achieves the best performance on dialogue-level RE, and comparable performance with the state-of-the-arts on sentence-level RE.\footnote{https://github.com/XueFuzhao/GDPNet}

\end{abstract}

%===========================
\section{Introduction}
\label{sec:intro}
%===========================

\noindent Given two entities and a piece of text where the two entities are mentioned in, the task of relation extraction (RE) is to predict the semantic relation between the two entities. The piece of text serves as the context for the prediction, which can be a short sentence, a long sentence, or even a dialog. 

We use an example from TACRED~\cite{zhang-etal-2017-position} to illustrate the RE task. In this example, we are interested in predicting the relation type between two entities: \textit{``Cathleen P. Black''} and \textit{``chairwoman''}. Based on a sentence: \textit{``Carey will succeed Cathleen P. Black, who held the position for 15 years and will take on a new role as chairwoman of Hearst Magazines, the company said''}, we aim to predict the two entities' relation to be ``per:title''. Note that, the relation types in RE tasks are predefined. For instance, TARCED defines 41 types and a special ``no relation'' type if a predicted relation is not covered in the 41 predefined types.

\begin{table}
\caption{An example from DialogRE dataset~\cite{yu-etal-2020-dialogue}.}
\vspace{-2mm}
\label{tbl-dialog-example}
\centering
\begin{tabular}{llll}
\toprule
\textbf{S1}: & \multicolumn{3}{l}{Hey Pheebs.}                                                                         \\
\textbf{S2}: & \multicolumn{3}{l}{Hey!}                                                                                \\
\textbf{S1}: & \multicolumn{3}{l}{Any sign of your \textbf{brother?}}                                                           \\
\textbf{S2}: & \multicolumn{3}{l}{No, but he is always late.}                                                          \\
\textbf{S1}: & \multicolumn{3}{l}{I thought you only met him once?}                                                    \\
\textbf{S2}: & \multicolumn{3}{l}{Yeah, I did. I think it sounds y’know big sistery,} \\
 & \multicolumn{3}{l}{ y’know,   ‘Frank’s always late.’} \\
\textbf{S1}: & \multicolumn{3}{l}{Well relax, he’ll be here.}                                                          \\
\midrule
    & \textbf{Argument Pair}                     & \textbf{Trigger}                    & \textbf{Relation type}                          \\
\textbf{R1}  & (Frank, S2)                       & brother                    & per:siblings                           \\
\textbf{R2}  & (S2, Pheebs)                      & none                       & per:alternate names                 \\
\bottomrule
\end{tabular}
%	\vskip -0.1in
\end{table}

Observe from the example, only a few words (\eg ``take a new role as'') in the given context are related to the semantic relation between the two entities.  Most of the remaining words in the given sentence are less relevant to the prediction. Another example from  DialogRE~\cite{yu-etal-2020-dialogue} dataset is shown in Table~\ref{tbl-dialog-example}.  Observe that Relation 1 (R1) can be easily predicted based on a trigger word (\eg ``brother''), despite the long conversation between S1 and S2. DialogRE even provides trigger word annotation, which is the smallest span of text that most clearly indicates the existence of the relation between two arguments (see Table~\ref{tbl-dialog-example}). This observation motives us to find and rely more on such indicative words for RE, particularly when the context is long.

Graph-based neural models have been widely adopted for RE due to their outstanding performance. Typically, each node in graph represents a token or an entity in the given text. There are multiple ways to construct edges. Many studies rely on an external parser converting text sequences to dependency trees to initialize the graph. Errors made by the parser therefore propagate to the graph. Recent studies directly learn a latent graph from  text~\cite{christopoulou-etal-2018-walk,christopoulou2019connecting,hashimoto-tsuruoka-2017-neural}. The challenge is to handle long texts, as in the example shown in Table~\ref{tbl-dialog-example}. It is difficult to learn latent graphs from long sequences, with token level node representations. \citet{christopoulou2019connecting} simplifies the latent graph by using predefined rules and extra labels, but these rules and labels are not readily available in raw data.

Similar to many other tasks, BERT-based models have demonstrated effectiveness on both sentence-level RE~\cite{wu2019enriching,joshi2020spanbert} and dialogue-level RE~\cite{yu-etal-2020-dialogue}. In BERT-based models, the ``[CLS]'' token is utilized as the task-specific representation for relation prediction. Although BERT can be regarded as a special case of a fully connected graph, it is too large and complex to be treated as a task-specific graph for relation extraction. More importantly, many words in the given context are less relevant to the relation prediction task.

In this paper, we propose a more general solution for constructing latent graphs without prior knowledge, for RE tasks. We start with a large latent graph initialized with all tokens/entities in the given context as nodes, based on their representations computed by BERT. Then we refine the graph through graph pooling operations with the aim of finding indicative words for relation extraction. In this sense, we focus on refining a task-specific graph on top of  BERT, by making full use of BERT token representations. We believe that a small-scale task-specific graph is critical for RE model to capture the relationships among indicative words that contain rich semantic information for relation prediction.  

%===========================
%===========================
%\citet{10.1145/2806416.2806502} is added here
%===========================
%===========================

When learning a latent graph, an edge between two tokens denotes their relationship abstracted from the text sequence. In RE tasks, the relationships between two tokens could be complex, including complicated syntactic relations and abstract semantic relations. Moreover, the relationships between two tokens are asymmetric in RE tasks. Following~\citet{10.1145/2806416.2806502}, we introduce the multi-view graph to fully capture different possible asymmetric relations between two arbitrary tokens. More specifically, we propose a Gaussian Graph Generator (GGG) to initialize the edges of the latent multi-view graph. In GGG, we first encode each node representation into multiple Gaussian distributions. Then the edge weights are computed by measuring the Kullback-Leibler (KL) divergence between the Gaussian distributions of different nodes. Due to the asymmetry of KL divergence, the graph generated by GGG is naturally a directed graph.

After initialization, the latent multi-view graph is very large, if the input sequence is long. It is difficult for the RE model to focus on the indicative  tokens for relation prediction. Thus, we propose a Dynamic Time Warping Pooling (DTWPool) to refine the graph, and to obtain hierarchical representations in an adaptive manner. By the regulation of SoftDTW~\cite{cuturi2017soft}, DTWPool refines the latent graph through a lower bound of pooling ratio, and reserves a flexible number of nodes in the multi-view graph. As a result, we obtain a task-specific graph with adaptive size to model the indicative tokens for relation extraction. Our contributions are summarized as follow:

\begin{itemize}
\item We propose a Gaussian Graph Generator (GGG) to initialize edges for latent multi-view graph by measuring KL divergence between different Gaussian distributions of tokens.
\item We propose a graph pooling method, DTWPool, to refine the latent multi-view graph learned from text sequence, with a flexible pooling ratio. To the best of our knowledge, this is the first work on multi-view graph pooling.
\item We combine GGG and DTWPool to form the GDPNet, and evaluate GDPNet on two benchmark datasets for RE. Experimental results demonstrate the effectiveness of GDPNet against SoTA baselines.

\end{itemize}

%===========================
\section{Related Work}
\label{sec:relate}
%===========================
We briefly review the related studies in four aspects, namely, RNN-based, graph-based, and BERT-based relation extraction methods, and graph pooling methods.

%===========================
\subsection{RNN-based Relation Extraction}
%===========================
Early works of relation extraction rely on hand-crafted features to represent pairs of entities~\cite{miwa2014modeling,gormley-etal-2015-improved}. Model effectiveness highly depends on the quality of hand-crafted features. Current works focus on learning based models, such as recurrent neural network (RNN) for RE. \citet{zhou-etal-2016-attention-based} propose bidirectional LSTM model to capture the long-term dependency between entity pairs. \citet{zhang-etal-2017-position} present PA-LSTM to encode global position information to boost the performance of RE.

%===========================
\subsection{Graph-based Relation Extraction}
%===========================
Graph-based models are now widely adopted in RE due to its effectiveness and strength in relational reasoning. \citet{zhang-etal-2018-graph} utilizes a graph convolutional network (GCN) to capture information over dependency structures. \citet{guo-etal-2019-attention} propose an attention guided GCN (AGGCN) to improve graph representations via self-attention mechanism.  AGGCN performs well on sentence-level RE, but it relies on external parser which may cause error propagation in graph generation. To alleviate error propagation, \citet{christopoulou-etal-2018-walk} and \citet{nan2020reasoning} propose to learn latent graph from text in an end-to-end manner, without the need of dependency trees generated by external parser. \citet{guo2020learning} treats the dependency structure as a latent variable and induces it from the unstructured text in an end-to-end fashion. In our model, we also generate a latent graph, but with two major differences. One is that our graph is a multi-view directed graph aiming to model all possible relationships between tokens. Second is that we focus on refining this multi-view graph to capture important words from long texts, for RE.

%===========================
\subsection{BERT-based Relation Extraction}
%===========================
Recently, large-scale pre-trained language models, such as BERT, have achieved SoTA performances on many tasks. Several works show that BERT-based models outperform both RNN and graph-based models with a large margin~\cite{wu2019enriching,joshi2020spanbert,yu-etal-2020-dialogue}. \citet{joshi2020spanbert} propose SpanBERT to learn better representations, and achieve SoTA performance on TACRED~\cite{zhang-etal-2017-position}, a sentence-level RE dataset. For dialogue-level RE task, \citet{yu-etal-2020-dialogue} present a BERTs model, which takes the speaker information into consideration and achieves the best result. In our solution, the multi-view graph is built on top of token representations by BERT.

\begin{figure*}
\centering
\includegraphics[width=0.9\textwidth]{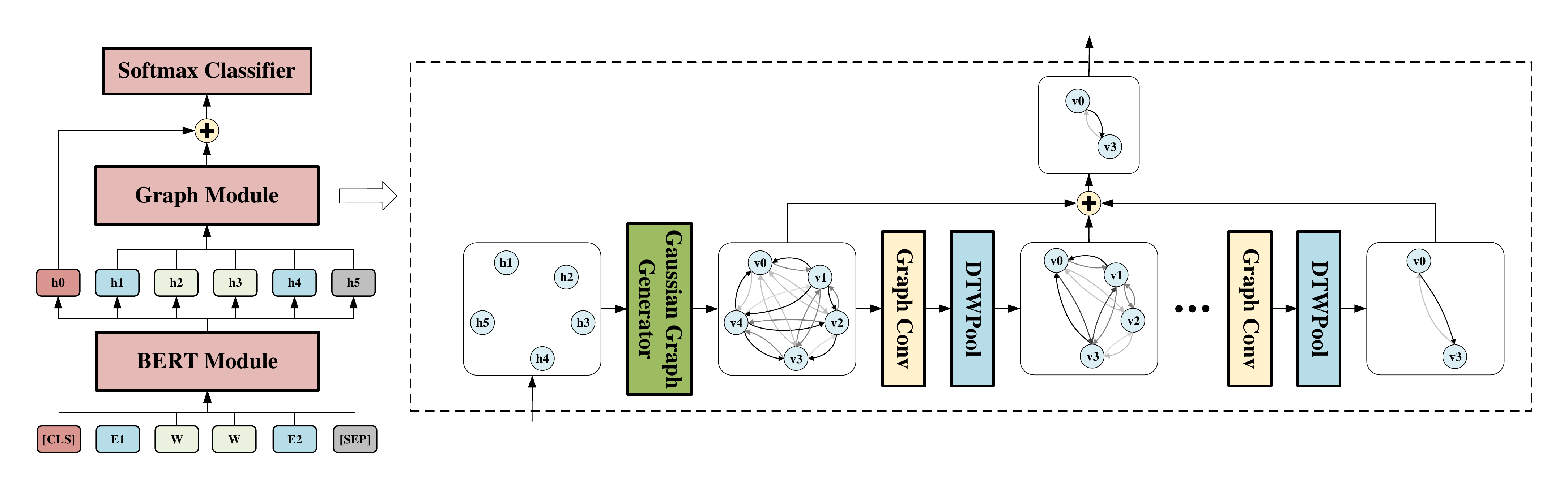}
\caption{The overall architecture of the proposed GDPNet. Entities $E_1$ and $E_2$ are  single-token entities in the illustration. }
\label{fig:overview}
\end{figure*}

%===========================
\subsection{Graph Pooling}
%===========================
Compared with graph convolution, graph pooling is another important but less investigated direction for learning graph. Graphs usually contain different substructures, and different nodes in a graph may play different roles. Hence, simply applying pooling operations like sum or average to encode the global node representations in graph may cause information loss~\cite{atwood2016diffusion,simonovsky2017dynamic}. To tackle this issue, several advanced graph pooling methods are proposed, including DiffPool~\cite{ying2018hierarchical}, TopKPool~\cite{gao2019graph}, SAGPool~\cite{pmlr-v97-lee19c} and StructPool~\cite{yuan2019structpool}. DiffPool generates a cluster assignment matrix over the nodes, to construct a differentiable pooling layer. TopKPool samples a subset of essential nodes, by manipulating a trainable projection vector. SAGPool further applies self-attention and graph convolution to improve TopKPool. StructPool proposes a graph pooling method based on conditional random fields, to improve the relation representations of different nodes. However, all these methods have not been evaluated on multi-view graphs, which can model complex relationships between nodes. In this paper, we propose DTWPool to process the latent multi-view graph learned from text sequence. Note that DTWPool is capable of utilizing adaptive pooling ratio instead of a fixed one.

%===========================
\section{Preliminary}
\label{sec:prelim}
%===========================

%===========================
\subsection{Problem Formulation}
%===========================
Let $X=\{x_1,x_2,\ldots,x_T\}$ be a sequence, where $x_t$ is the $t^{th}$ token in the sequence, and $T$ is the number of tokens. For sentence-level RE, $X$ denotes the given sentence. For dialogue-level RE, $X$ represents the entire dialogue. That is, in our problem formulation, we do not explicitly distinguish sentence and dialogue. To predict relations, we are given two entities, subject entity $X_s$ and object entity $X_o$. Both $X_s$ and $X_o$ are sub-sequences of $X$. An entity may contain one or more tokens, \eg  $X_s=\{x_{s},x_{s+1},\ldots,x_{s+m-1}\}$ where $s$ denotes the starting position of $X_s$ in sequence $X$ and $m$ is the number of tokens in $X_s$.  Given $X$, $X_s$, and $X_o$, the goal of relation extraction is to predict the relation $r\in R$ between $X_s$ and $X_o$, where $R$ is a set of predefined relation types in the dataset.

\subsection{Multi-view Graph}
In a multi-view graph, there exist multiple edges between a pair of nodes, each edge from one view. Formally, we can represent a multi-view graph as $G=(V,A^1,A^2,\ldots,A^N)$, where $V$ is the set of nodes, $A$ is an adjacent matrix, and  $N$ is the number of views. 

In our model, we aim to use multi-view graph to model the complex (\eg syntactic and semantic) relationships between tokens in the given sequence. In this multi-view graph,  each token in sequence $X$ corresponds to one node. Note that an entity may contain multiple tokens, and there are multiple ways of handling entity tokens (\ie tokens in $X_s$ and $X_o$). For easy presentation, we treat tokens in entities the same as other tokens in the sequence in following discussions.

%===========================
\section{Methodology}
\label{sec:method}
%===========================
The overall architecture of GDPNet is shown on the left-hand side in Figure~\ref{fig:overview}. There are three key components: BERT module, graph module, and SoftMax classifier. The BERT module encodes tokens into the corresponding feature representations. Illustrated on the right-hand side of Figure~\ref{fig:overview}, the graph module takes in token representations from BERT and constructs a multi-view graph with a Gaussian Graph Generator (GGG). Then the graph is refined through multiple interactions of graph convolution and DTWPool. Finally, the refined latent graph is fed into the SoftMax classifier to predict relation type.

%===========================
\subsection{BERT Module}
%===========================

We utilize BERT as the feature encoder to extract token representations due to its effectiveness in representation learning~\cite{joshi2020spanbert,yu-etal-2020-dialogue}. Given a sequence $X$ with $T$ tokens, we map $X$ to a BERT input sequence $X_{input}=\{x_0,x_1,x_2,\ldots,x_T,x_{T+1}\}$. Here, $x_0$ denotes the ``[CLS]'' token which represents the start of sequence $X$, and $x_{T+1}$ is the ``[SEP]'' token which represents the end of the sequence. The corresponding token representations from BERT are denoted by  $H=\{h_0,h_1,h_2,\ldots,h_T,h_{T+1}\}$. Existing BERT-based solutions for RE only take $h_0$, \ie the representation of ``[CLS]'' token, as the input of SoftMax classifier to predict the relation type~\cite{joshi2020spanbert,yu-etal-2020-dialogue}. In GDPNet, we fully utilize the entire token representations $H$ through the graph module. To be detailed shortly, the graph module learns a task-specific graph using tokens $\{h_1,h_2,\ldots,h_T,h_{T+1}\}$. The learned graph is then combined with $h_0$ to be the input to the SoftMax classifier.

\subsection{Graph Module}
The graph module consists of Gaussian Graph Generator (GGG), multiple layers of graph convolution and DTWPool. The GGG is designed to generate the latent multi-view graph, while the graph convolution and DTWPool layers are applied for graph refinement.

%===========================
\subsubsection{Gaussian Graph Generator}
%===========================
The output representations of BERT module are divided into two parts. $h_0$ for token ``[CLS]'' is considered as the task-specific token of the entire sequence, which is the first part. The remaining representations $\{h_1,h_2,...,h_T,h_{T+1}\}$ form the second part. We generate a multi-view graph from the second part to model the relationships between tokens. 

%AIXIN: Up to this point

We denote the initial node representations of the latent graph as $V^0=\{v^0_1,v^0_2,\ldots,v^0_{T+1}\}$, where each node corresponds to a token representation. Then, we propose a Gaussian Graph Generator (GGG) to initialize the edges of the latent multi-view graph, based on $V^0$. 
Specifically, we first encode each node $v^0_i$ into multiple Gaussian distributions as:

\begin{equation}
\begin{aligned}
   \{\mu^1_i,\mu^2_i,\ldots,\mu^N_i\} & = g_\theta(v^0_i) \\
   \{\sigma^1_i,\sigma^2_i,\ldots,\sigma^N_i\} & = \phi \left(g'_\theta(v^0_i)\right)
\end{aligned}
\end{equation}
where $g_\theta$ and $g'_\theta$ are two trainable neural networks, $\phi$ is a non-linear activation function and $N$ denotes the number of views in the multi-view graph. We set the activation function $\phi$ as the SoftPlus function, since the standard deviation of Gaussian distribution is bounded on $(0,+\infty)$. Consequently, we obtain a number of Gaussian distributions $\{\mathcal{N}^n_1,\mathcal{N}^n_2,\ldots,\mathcal{N}^n_{T+1}\}$ for the $n^{th}$ view of the multi-view graph. Each Gaussian distribution here $\mathcal{N}^n_i(\mu^n_i,{\sigma^n_i}^2)$ corresponds to a node representation $v^0_i$.

The purpose of the multi-view graph is to capture all possible relations between tokens, so we encourage message propagation between token representations with large semantic differences. We adopt KL divergence between the Gaussian distributions of two tokens to model edge weight. Specifically, edge weight between $i^{th}$ node and $j^{th}$ node on the $n^{th}$ view is computed as:
\begin{equation}
    e^n_{ij} = \mathrm{KL}\left(\mathcal{N}^n_i(\mu^n_i,{\sigma^n_i}^2) || \mathcal{N}^n_j(\mu^n_j,{\sigma^n_j}^2)\right )
\end{equation}
After computing edges between nodes on each view, we obtain multiple adjacent matrices $\{A^1,A^2,...,A^N\}$, one for each view. Thus, the generated multi-view graph is written as $G=(V^0,A^1,A^2,...,A^N)$. Due to the asymmetry nature of KL divergence, $G$ is a directed multi-view graph.

%===========================
\subsubsection{Multi-view Graph Convolution}
%===========================
Inspired by \citet{guo-etal-2019-attention}, we further employ multi-view graph convolution with dense connections~\cite{guo2019densely} to capture structural information on the graphs. With the usage of dense connections, we can train a deeper model to capture both local and non-local information. The multi-view graph convolution is written as:
\begin{equation}
    v^{(\ell)}_{n_i} = \rho \left(\sum_{j=1}^{T}A^n_{ij}W^{(\ell)}_n k^{(\ell)}_j+b^{(\ell)}_n\right )
\end{equation}
where $W^{(\ell)}_n$ and $b^{(\ell)}_n$ are the trainable weight and bias of the $n^{th}$ view, respectively. $\rho$ denotes an activation function and $k^{(\ell)}_j$ is the concatenation of the initial node representation and the node representations produced in sub-layers $1,...,\ell-1$. The output of the first graph convolution layer is $V^1=\{v^1_1,v^1_2,\ldots,v^1_{T+1}\}=\{k^{(\ell)}_1,k^{(\ell)}_2,\ldots,k^{(\ell)}_{T+1}\}$. We  refer readers to the Densely Connected Graph Convolutional Network~\cite{guo2019densely} for more details.

\begin{table*}
\centering
\caption{Performance of all models on DialogRE.  $\sigma$ denotes the standard deviation computed from five runs of each model.}
\label{tbl-dialog_eval}
\begin{tabular}{l|cc|cc}
\toprule
\multirow{2}*{Model} & \multicolumn{2}{c|}{Dev set} & \multicolumn{2}{c}{Test set} \\
 & $F1$ ($\sigma$)         & $F1c$ ($\sigma$)        & $F1$ ($\sigma$)          & $F1c$ ($\sigma$)        \\ 
\midrule
CNN \cite{lawrence1997face}         & 46.1 (0.7)  & 43.7 (0.5)  & 48.0 (1.5)   & 45.0 (1.4)  \\
LSTM \cite{hochreiter1997long}      & 46.7 (1.1)  & 44.2 (0.8)  & 47.4 (0.6)   & 44.9 (0.7)  \\
BiLSTM \cite{graves2005framewise}   & 48.1 (1.0)  & 44.3 (1.3)  & 48.6 (1.0)   & 45.0 (1.3)  \\
BERT \cite{devlin-etal-2019-bert}   & 60.6 (1.2)  & 55.4 (0.9)  & 58.5 (2.0)   & 53.2 (1.6)  \\
BERTs \cite{yu-etal-2020-dialogue}  & 63.0 (1.5)  & 57.3 (1.2)  & 61.2 (0.9)   & 55.4 (0.9)  \\
GDPNet (our model)                              & \textbf{67.1} (1.0) & \textbf{61.5} (0.8) & \textbf{64.9} (1.1) & \textbf{60.1} (0.9)      \\
\bottomrule
\end{tabular}
\end{table*}

\begin{table}
\caption{Text of the first dialogue in test set of DialogRE. The tokens in bold are selected by GDPNet.}
\label{tbl-dialog-case_study}
\centering
\begin{tabular}{llll}
\toprule
S1: & \multicolumn{3}{l}{Hey, you guys! Look \textbf{what I} found! Look at this!} \\
    & \multicolumn{3}{l}{\textbf{That’s my} Mom’s \textbf{writing!} Look.}    \\
S2: & \multicolumn{3}{l}{Me \textbf{and Frank} and Phoebe, Graduation 1965.}  \\
S1: & \multicolumn{3}{l}{Y'know what that \textbf{means}?}   \\
S3: & \multicolumn{3}{l}{That you’re actually 50?}               \\
S1: & \multicolumn{3}{l}{No-no, that’s \textbf{not}, that’s not me Phoebe, \textbf{that’s} her}     \\
    & \multicolumn{3}{l}{pal Phoebe. According to \textbf{her} high \textbf{school} year-}        \\
    & \multicolumn{3}{l}{book, they were \textbf{like} \textbf{B.F}.F. Best \textbf{Friends} Forever.} \\

\midrule
    & Argument Pair  & Trigger   & Relation type \\
R1  & (S1, Frank)         & high school  & per:alumni     \\
&            & yearbook                    &                         \\
\bottomrule
\end{tabular}
\end{table}

%===========================
\subsubsection{Dynamic Time Warping Pooling} 
%===========================
After graph convolution updates node representations by message propagation, a Dynamic Time Warping Pooling (DTWPool) is introduced to refine the latent multi-view graph. In DTWPool, we first refer to SAGPool~\cite{pmlr-v97-lee19c} to calculate the attention scores on each view of the graph:
\begin{equation}
    s_{n_i} = \alpha \left(\sum_{j=1}^{T}A^n_{ij}W_{pool} v_j+b_{pool}\right)
\end{equation}
where $W_{pool}$ and $b_{pool}$ are trainable weight and bias of the pooling operation, respectively. $\alpha$ denotes an activation function and $s_{n_i}$ is the attention weight before the SoftMax activation. For $n^{th}$ view of the latent multi-view graph, we obtain a score set $S_n=\{s_{n_1},s_{n_2},\ldots,s_{n_{T+1}}\}$. We keep the node selection method of SAGPool to retain a portion of nodes in the input graph even when the sizes and structures of the graphs are varied. After  node selection, the retained nodes of $n^{th}$ view are a subset of the $V^1$, \eg $V^2_n=\{v^2_1,v^2_5,...,v^2_T\}$. As our latent graph has multiple views, we can derive different subsets from $V^1$ from different views. 

Existing graph pooling approaches, \eg SAGPool, only allow a fixed ratio for node pooling. Due to the nature of multi-view graph, DTWPool refines the graph adaptively by getting the union set of nodes from different views:

\begin{equation}
    V^2 = V^2_1\cup V^2_2\cup\ldots \cup V^2_N
\end{equation}
where $V^2$ is the union set of the subsets selected from all different views. If we set a fixed pooling ratio $r\in[0,1]$ on each view, the pooling ratio of DTWPool, \eg ratio of the number of nodes in $V^2$ to the number of nodes in $V^1$, could be a flexible decimal $r_{real}\in[r,1]$. We operate the graph convolution and DTWPool iteratively in the graph module, so we have a sequence of graphs $\{G^1,G^2,\ldots,G^D\}$, where $D$ is the number of graph pooling layers.

The number of informative nodes varies in different text sequences. It is important to preserve important information along the process of graph pooling. The nodes in this graph embed rich context information, so it would be beneficial to summarize the context into the pooled nodes. To this end, we propose to adopt SoftDTW to guide the graph pooling. SoftDTW is a differentiable loss function, designed for finding the best possible alignment between two sequences with different lengths~\cite{cuturi2017soft}.  % which can be defined as: 
\begin{equation}
    \mathrm{DTW}_\gamma(L_1,L_2)=\mathrm{min}^{\gamma}\left \{ \langle M, \Delta(L_1,L_2) \rangle, M \in \mathcal{M}\right \}
\end{equation}
Here, $L_1$ and $L_2$ are two sequences of different lengths, $\Delta(L_1,L_2)$ is the cost matrix, and $\mathcal{M}$ is a set of binary alignment matrices.

In GDPNet, we use the SoftDTW loss to minimize the distance between the original graph and the last pooled graph:
\begin{equation}
\label{learning_func}
    \mathcal{L}=\mathrm{CSE}(r,\hat{r})+\lambda \mathrm{{DTW}}_\gamma(V^1,V^D)
\end{equation}
where $\mathcal{L}$ denotes the overall training objective, $\mathrm{CSE}$ is the Cross Entropy loss function, $\lambda$ is a hyper-parameter to balance the contribution of $\mathrm{DTW}$, and $V^D$ is the nodes in the graph after the last DTWPool layer. Because SoftDTW loss is designed for aligning two sequences, we employ it to encourage the nodes after pooling to cover more local context representations. With SoftDTM loss, DTWPool is guided to refine the graph without losing much context information.

To minimize information loss, we concatenate the node representations of the intermediate graphs created during the pooling process to derive the final graph $V$, similar to learning graph dense connections~\cite{guo2019densely}. As our pooled graphs have different sizes, we only concatenate the node representations in $\{V^2,V^4,...,V^D\}$ for all nodes that are included in graph $V^D$. Thus, the number of nodes in the final graph $V$ is the same as that in $V^D$. 

%===========================
\subsubsection{Classifier}
%===========================
Given the final graph $V$, we adopt a neural network with max-pooling to compute the representation of the graph. The computed representation is then concatenated with the representation of ``[CLS]'' token $h_0$ to form the final representation.
\begin{equation}
    h_{final}=[h_0;f(V)]
\end{equation}
Here, $f$ is a neural network with max-pooling, which maps the $V \in \mathbb{R}^{Q \times T'}$ to $f(V) \in \mathbb{R}^{q \times 1}$, $Q=D*q$. $D$ is the number of graph pooling layers in graph module, $q$ is the dimension of token representation, and $T'$ is the number of nodes in $V$.

%===========================
\section{Experiments}
%===========================
Our proposed GDPNet can be applied to both sentence-level and dialogue-level RE tasks. Due to the differences in data formats, applicable baseline models, and the way in handling subject and object entities $X_s$ and $X_o$, we conduct two sets of experiments, comparing GDPNets to SoTA models on the two tasks. We also show how GDPNet can be easily modified to achieve a fair comparison with SoTA models on each task.   

%===========================
\subsection{Dialogue-level Relation Extraction}
%===========================

DialogRE is the first human-annotated dialogue-level RE dataset~\cite{yu-etal-2020-dialogue}. It contains $1,788$ dialogues originating from the complete transcripts of a famous American television situation comedy. There are 36 relation types predefined in DialogRE. An example is given in Table~\ref{tbl-dialog-example}. 

%===========================
\subsubsection{Baseline Models and Experimental Setup}
%===========================
We evaluate GDPNet against the recently proposed BERTs~\cite{yu-etal-2020-dialogue}. BERTs is a speaker-aware modification of BERT, and achieves best performance on dialogue-level RE. For the completeness of experiments, we also include popular baseline models: CNN, LSTM, BiLSTM and BERT models.

For fair comparison, we use the same input format and hyperparameter settings as in BERTs. Specifically, the given $X$, $X_s$, and $X_o$, are concatenated with classification token [CLS] and separator token [SEP] to form an input sequence [CLS]$X$[SEP]$X_s$[SEP]$X_o$[SEP]. All token representations except [CLS] are fed into our graph module. To incorporate speaker information, for the sentences that contain $X_s$ or $X_o$, the text indicating speaker \eg ``Speaker 1'', is replaced by a specific token, [$S_1$] or [$S_2$]. Note that the trigger words are treated as normal tokens. Adam~\cite{Kingma2015AdamAM} with learning rate of $3e-5$ is employed and the lower bound of pooling ratio is set to $0.7$. We use 3 DTWPool layers. As graph pooling operation is performed in each DTWPool layer, only a few nodes are included in the final graph. We use both $F1$ and $F1c$ scores as the evaluation metrics. $F1c$ is proposed by \citet{yu-etal-2020-dialogue}, and it is computed by only taking in the early part of a dialogue as input, instead of the entire dialogue.

%===========================
\subsubsection{Results on DialogRE}
%===========================
Table~\ref{tbl-dialog_eval} summarizes the results on DialogRE. Observe that BERT-based models significantly outperform CNN and LSTM-based models. BERTs is superior to BERT because BERTs incorporates speaker-related information. Following the same input format and setting, GDPNet is built on top of BERTs. That is, BERTs acts as the BERT module for feature extraction in GDPNet (see Figure~\ref{fig:overview}). Shown in Table~\ref{tbl-dialog_eval},  GDPNet outperforms BERTs by $3.7$ and $4.7$ points in $F1$ and $F1c$, respectively, on test set.

\begin{figure}
\centering
\includegraphics[width=0.85\columnwidth]{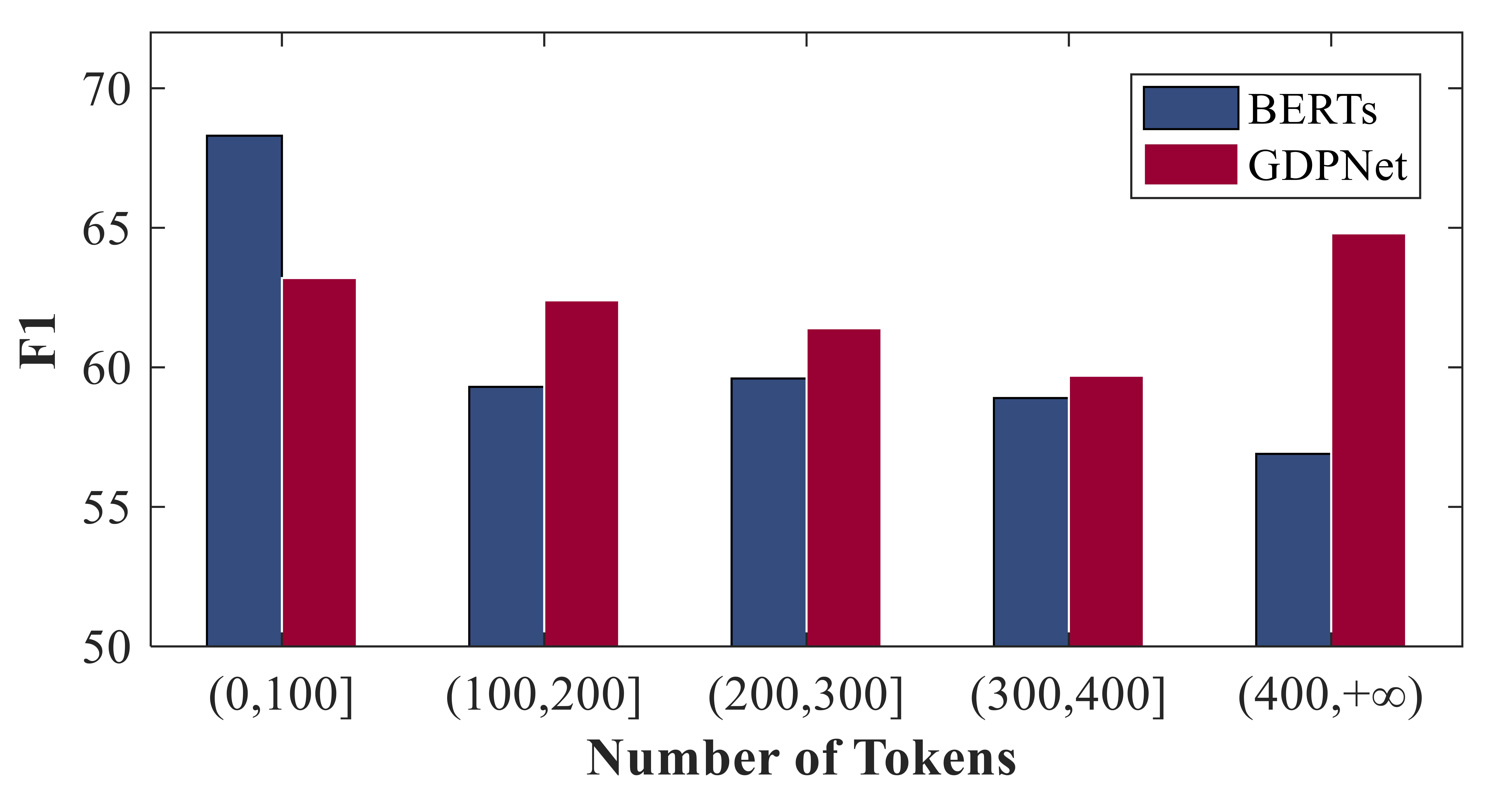}
\caption{GDPNet and BERTs on different dialogue lengths.}
\label{fig:GDPvsBERTs}
\end{figure}

GDPNet is designed to find key information from long sequences for effective RE.  Thus, we expect that GDPNet is capable of tackling long sequences better.  We group the dialogues in DialogRE test set into five subsets by their length, \ie number of tokens. Figure~\ref{fig:GDPvsBERTs} reports $F1$ scores of GDPNet and BERTs on the five subsets. GDPNet consistently outperforms BERTs when dialogue length is more than 100 tokens. In particular, GDPNet surpasses BERTs by a large margin on the dialogues with over 400 tokens. This comparison shows that GDPNet is effective in modeling long sequences through refining the latent graph built on top of token representations.

As an example, Table~\ref{tbl-dialog-case_study} shows the tokens selected by GDPNet on the first dialogue in the test set of DialogRE. Our model selects informative tokens like ``her'' "school" to predict the ``per:alumn'' relation. Some less related tokens, like ``you're actually 50?'', are ignored by our model.

\begin{table}
\small
\centering
\caption{An ablation study on GDPNet model. For models without GGG, we replaced GGG by multi-head attention to initialize the edges of the multi-view graph. For the model without DTWloss, we remove the second term in Equation~\ref{learning_func}. All DTWPool layers are removed for the models without DTWPool.}
\label{tbl-dialogre_a_study}
\begin{tabular}{l|c c}
\toprule
Model                 & $F1$($\sigma$)        & $F1c$($\sigma$)       \\ 
\midrule
GDPNet                & \textbf{64.9} (1.1) & \textbf{60.1} (0.9) \\
 with Homogeneous GGG  & 63.5 (0.7)          & 58.4 (0.6) \\
 w/o GGG               & 62.1 (1.6)          & 58.1 (1.1) \\
 w/o DTWloss           & 63.4 (1.4)          & 58.6 (1.3) \\
 w/o DTWPool           & 48.9 (1.1)          & 22.4 (1.0) \\
 w/o GGG \& DTWPool   & 48.2 (1.4)          & 21.8 (1.0) \\ 
\bottomrule
\end{tabular}
\end{table}

%===========================
\subsubsection{Ablation Study}
%===========================
We conduct ablative experiments on DialogRE to evaluate the effectiveness of the two main components in GDPNet, \ie Gaussian Graph Generator and Dynamic Time Warping Pooling. The results are reported in Table~\ref{tbl-dialogre_a_study}. 

We first replace the multi-view graph by a simple homogeneous graph, which is the same as setting the number of views to one. The performance degradation suggests that multi-view graph is beneficial as it models complex relationships among tokens. Next, we evaluate GGG by replacing GGG with multi-head attention. The results show that without GGG, performance drops. GGG initializes the latent graph by measuring the difference between two Gaussian distributions generated from node representations, which decouples the dependency between token representations and graph edges. To evaluate the impact of DTWPool, we first drop DTWloss by removing the second term in Equation~\ref{learning_func}, and there is a slight performance drop. Without DTWloss, DTWPool degenerates into the multi-view version of SAGPool~\cite{pmlr-v97-lee19c}. This result indicates that DTWPool outperforms SAGPool when tackling the multi-view graph learned from token sequence. When all the DTWPool layers are removed, the performance of GDPNet decreases dramatically, which shows that DTWPool is crucial for the GDPNet. After removing both GGG and DTWPool, the performance of GDPNet is even worse. To summarize, DTWPool is crucial for learning a task-specific graph from a large latent multi-view graph. The final graph learned effectively filters out less useful information from a long sequence for effective relation extraction.

%===========================
\subsection{Sentence-level Relation Extraction}
%===========================

We evaluate GDPNet for sentence-level RE on two datasets TACRED~\cite{zhang-etal-2017-position} and TACRED-Revisit~\cite{alt2020tacred}. TACRED is a widely used large-scale sentence-level relation extraction dataset. It contains more than 106K sentences drawn from the yearly TACKBP4 challenge, and 42 different relations (41 common relation types and a special ``no relation'' type). The subject mentions in TACRED are person and organization, while object mentions are in 16 fine-grained types, including date, location, etc. The TACRED-Revisit dataset, released recently, corrects the wrong labels in the development and test sets of TACRED.

\begin{table*}
\small
\centering
\caption{Performance of all models on TACRED and TACRED-Revisit.  For the models without reported performance on TACRED-Revisit, we run the released code if available, and mark results obtained by asterisk(*). We also run the released code of SpanBERT on TACRED-Revisit,  and we obtain the same results as reported in~\cite{alt2020tacred}.}
\label{tbl-tacred_eval}
\begin{tabular}{l|lll|lll}
\toprule
\multirow{2}*{Model} & \multicolumn{3}{c|}{TACRED} & \multicolumn{3}{c}{TACRED-Revisit} \\
& $Pr$       & $Re$       & $F1$     & $Pr$          & $Re$          & $F1$        \\ 
\midrule
LSTM~\cite{zhang-etal-2017-position}    & 65.7 & 59.9 & 62.7 & 71.5* & 69.7* & 70.6* \\
PA-LSTM~\cite{zhang-etal-2017-position} & 65.7 & 64.5 & 65.1 & 74.5* & 74.1* & 74.3* \\
C-AGGCN~\cite{guo-etal-2019-attention}  & 73.1 & 60.9 & 68.2 & 77.7* & 73.4* & 75.5* \\
LST-AGCN~\cite{sun2020relation}         & -    & -    & 68.8 & -    & -    & -    \\ 
\midrule
SpanBERT~\cite{joshi2020spanbert}       & 70.8 & 70.9 & \textbf{70.8} & 75.7* & 80.7* & 78.0* \\
GDPNet (Our model)                                  & 72.0 & 69.0 & 70.5 & 79.4 & 81.0 & \textbf{80.2} \\
\midrule
KnowBERT~\cite{Peters2019KnowledgeEC}   & 71.6 & 71.4 & \textbf{71.5} & -  & - & 79.3  \\
\bottomrule
\end{tabular}
\end{table*}

\begin{table}
\small
\centering
\caption{Percentage (\%) of the tokens selected in the final graph from sequence, \eg an entire dialogue in DialogRE or the whole sentence in TACRED. Non-repetitive tokens are the tokens that appear only once in the sequence; repetitive tokens appear two or more times in the sequence. Trigger tokens are key tokens annotated in DialogRE.}
\label{tbl-q_study}
\begin{tabular}{l| r r}
\toprule
 Type of tokens    & DialogRE & TACRED \\
\midrule
All tokens          & 15.6     & 66.3   \\
Non-repetitive tokens & 23.5     & 67.6   \\
Repetitive tokens    & 10.0      & 58.1  \\
Trigger tokens    & 32.1      & -  \\
\bottomrule
\end{tabular}
\end{table}

%===========================
\subsubsection{Baseline Models and Experimental Setup}
%===========================
To the best of our knowledge, SpanBERT~\cite{joshi2020spanbert} is the best performing sentence-level RE model without incorporating any external knowledge and parser. We consider SpanBERT as a strong baseline to benchmark our GDPNet. We also include RNN- and graph-based models. Meanwhile, we report the results of KnowBERT, which incorporates external resources for training~\cite{Peters2019KnowledgeEC}.

We use the same input format and hyperparameter settings as in SpanBERT. Subject entity $X_s$ and object entity $X_o$ are each replaced by a sequence of ``[SUBJ-NER]'' or ``[OBJ-NER]'' tokens. Then [CLS]$X$[SEP] forms the input to the models. This is different from the settings in DialogRE where  $X_s$ and $X_o$ are appended to the input sequence $X$. Parameter optimization is again performed by Adam~\cite{Kingma2015AdamAM} with learning rate of $2e-5$. Since the sequence length in TACRED is much shorter than that in DialogRE, we set the lower bound of pooling ratio to $0.8$.

%===========================
\subsubsection{Results on TACRED}
%===========================
The results on TACRED and TACRED-Revisit are summarized in Table~\ref{tbl-tacred_eval}. Similar observations hold, that BERT-based models (\ie SpanBERT and KnowBERT), significantly outperform non-BERT models (\ie LSTM, PA-LSTM, C-AGGCN and LST-AGCN) on both versions of TACRED. 

GDPNet achieves comparable performance with SpanBERT\footnote{A very recent study~\cite{chen2020efficient} also combines SpanBERT with GCN, but it still relies on external parser for graph generation. In contrast, our GDPNet regards the graph as latent variable, which is more general and feasible.} on TACRED, and better results on TACRED-Revisit. Compared to KnowBERT, which utilize external knowledge in its training, GDPNet's $F1$ is lower by 1 point on TACRED, but is higher by almost 1 point on TACRED-Revisit. 

Compared to dialogue, sentence is much shorter and BERT-based models are effective in capturing the key information. As GDPNet is designed for handling long sequences, we do not expect it to outperform SoTA models, but GDPNet remains competitive for sentence-level RE.

%===========================
\subsection{Quantitative Analysis}
%===========================
Our last experiment is to analyze DTWPool in GDPNet. DTWPool aims to identify indicative tokens for relation extraction through refining the latent multi-view graph. Table~\ref{tbl-q_study} reports the percentage of tokens selected in the final graph after the DTWPool process, on both DialogRE and TACRED datasets. We separate the repetitive tokens and non-repetitive tokens based on the original input, \ie whether the word appears only once or multiple times in the input sequence. Repetitive tokens, in general, define the topic of the sentence or dialogue. However, the relation type between two entities is seldom described repetitively. In fact, given the same dialogue, we may predict different relation types between different pairs of entities. With this in mind, we consider repetitive tokens are less important compared to non-repetitive tokens for RE tasks in general. Shown in Table~\ref{tbl-q_study}, DTWPool selects more non-repetitive tokens than repetitive tokens on both datasets, in particular, on the DialogRE dataset. More interestingly, DialogRE provides manually annotated trigger tokens that are indicative to the relation type. DTWPool selects 32.1\% of trigger tokens, given that only 15.6\% of tokens are selected among all tokens. That is, trigger tokens are selected with a much higher chance than random. This analysis shows that DTWPool is capable of selecting indicative tokens for relation extraction.

%===========================
\section{Conclusion}
%===========================

In this paper, we propose GDPNet for relation extraction. GDPNet is designed to find indicative words from long sequences (\eg dialogues) for effective relation extraction. We show that GDPNet achieves the best performance on dialogue-level RE. In particular, GDPNet achieves much better performance than BERT-based models when the dialogue is long. The key of the GDPNet is to construct a latent multi-view graph to model possible relationships among tokens in a long sequence, and then to refine the graph by DTWPool. From the results on DialogRE and TACRED, we show there is a great potential of this mechanism in dealing with long sequences. To evaluate the effectiveness of this mechanism on other tasks is part of our future work.

%===========================
\section{Acknowledgments}
%===========================
Aixin Sun is supported by the Agency for Science, Technology and Research (A*STAR) AME Programmatic Fund (Grant No. A19E2b0098). Hao Zhang is supported by A*STAR AME Programmatic Funds (Grant No. A18A1b0045 and A18A2b0046).

%===========================
\section{Appendix}
%===========================

\subsection{Software Packages and Hardware Specification}
The GDPNet is implemented by using PyTorch 1.4 with CUDA 10.1. Our implementation also uses the SoftDTW\footnote{\url{https://github.com/Maghoumi/pytorch-softdtw-cuda}} toolkit. All experiments are conducted on a desktop with Intel i7-8750H CPU, DDR4 16GB memory, and a single NVIDIA GeForce RTX 1070 GPU. We also reproduced our results on Quadro RTX 8000 GPU.

\subsection{Hyper-Parameter Settings}

The hyper-parameter settings on the two datasets, DialogRE and TACRED, are listed as follows. 

\begin{center}
\begin{tabular}{l|l l}
\toprule
Parameter                  & DialogRE & TACRED   \\ \midrule
Epoch                     & 20       & 10       \\
Batch Size                 & 24       & 32       \\
Learning rate             & 3e-5     & 2e-5 \\
Dropout                   & 0.5      & 0.5      \\
Hidden units of graph     & 300      & 300      \\
Number of views           & 3        & 3        \\
Number of DTWPool layers  & 3        & 3        \\
Pooling ratio lower bound & 0.7      & 0.8      \\
Weight of SoftDTW loss    & 1e-6     & 2e-4    \\
\bottomrule
\end{tabular}
\end{center}

We set hyper-parameters (epoch, batch size, learning rate and dropout rate) of the backbone model the same as the corresponding SoTA models for fair comparison, i.e., BERTs~\cite{yu-etal-2020-dialogue} for DialogRE, and SpanBERT~\cite{joshi2020spanbert} for TACRED. 

The hidden units of graph, number of views, and number of DTWPool layers are set  according to AGGCN~\cite{guo-etal-2019-attention}. Although AGGCN does not contain graph pooling operation, we use the same number of GCN layers and DTWPool layers in our evaluation. 

We set a higher pooling ratio for the TACRED dataset due to its shorter sequence length compared to DialogRE. Consequently, the ratio of nodes retained in the final graph of TACRED is higher than that of DialogRE, which leads to lower SoftDTW loss for TACRED compared to DialogRE. For this reason, we set a larger weight of SoftDTW loss for TACRED.

\bibliography{reference}

\end{document}